\documentclass[journal]{IEEEtran}
\usepackage{graphicx}
\usepackage{indentfirst,amssymb,amsmath}
\usepackage{amsthm}
\usepackage{CJK}
\usepackage{multirow}
\usepackage{epstopdf}
\usepackage{algorithm}
\usepackage{algorithmic}
\usepackage{subfigure}
\usepackage{caption}
\usepackage{stfloats}
\usepackage{xspace}

\makeatletter
\DeclareRobustCommand\onedot{\futurelet\@let@token\@onedot}
\def\@onedot{\ifx\@let@token.\else.\null\fi\xspace}

\def\eg{\emph{e.g}\onedot} 
\def\ie{\emph{i.e}\onedot}

\def\etal{\emph{et al}\onedot}
\makeatother
\begin{document}
%
\title{Radio-Assisted Human Detection}
%
%
%

\author{Chengrun Qiu, Dongheng Zhang, Yang Hu, Houqiang Li,~\IEEEmembership{Fellow,~IEEE,} Qibin Sun,~\IEEEmembership{Fellow,~IEEE,}
        and Yan Chen,~\IEEEmembership{Senior Member,~IEEE}
\thanks{Chengrun Qiu and Dongheng Zhang are with School of Information and Communication Engineering, University of Electronic Science and Technology of China, E-mail: \{cr$\_$qiu, eedhzhang\}@std.uestc.edu.cn.}
\thanks{Yang Hu and Houqiang Li are with School of Information Science and Technology, University of Science and Technology of China, Email:\{eeyhu,lihq\}@ustc.edu.cn.}
\thanks{Qibin Sun and Yan Chen are with School of Cyberspace Security, University of Science and Technology of China, Email:\{qibinsun,eecyan\}@ustc.edu.cn.}
}

\maketitle

\begin{abstract}
In this paper, we propose a radio-assisted human detection framework by incorporating radio information into the state-of-the-art detection methods, including anchor-based one-stage detectors and two-stage detectors. We extract the radio localization and identifer information from the radio signals to assist the human detection, due to which the problem of false positives and false negatives can be greatly alleviated.
For both detectors, we use the confidence score revision based on the radio localization to improve the detection performance.
For two-stage detection methods, we propose to utilize the region proposals generated from radio localization rather than relying on region proposal network (RPN).
Moreover, with the radio identifier information, a non-max suppression method with the radio localization constraint has also been proposed to further suppress the false detections and reduce miss detections. Experiments on the simulative Microsoft COCO dataset and Caltech pedestrian datasets show that the mean average precision (mAP) and the miss rate of the state-of-the-art detection methods can be improved with the aid of radio information. Finally, we conduct experiments in real-world scenarios to demonstrate the feasibility of our proposed method in practice.
\end{abstract}


%
\IEEEpeerreviewmaketitle

\section{Introduction}

Human detection is a fundamental problem in computer vision, which can power many other vision tasks such as instance segmentation \cite{MaskRCNN,segmentation} and pose estimation \cite{Pose1,Pose2}. Generally, existing detection methods first extract the features from the images and then output the bounding boxes together with the corresponding confidence scores. Compared with the traditional methods \cite{HoG, VJ}, deep learning based detection methods can achieve much better performance \cite{MaskRCNN,YOLO9000} since the deep neural backbone can extract more useful features from the images.

\begin{figure}
  \centering
  \includegraphics[width=1\linewidth]{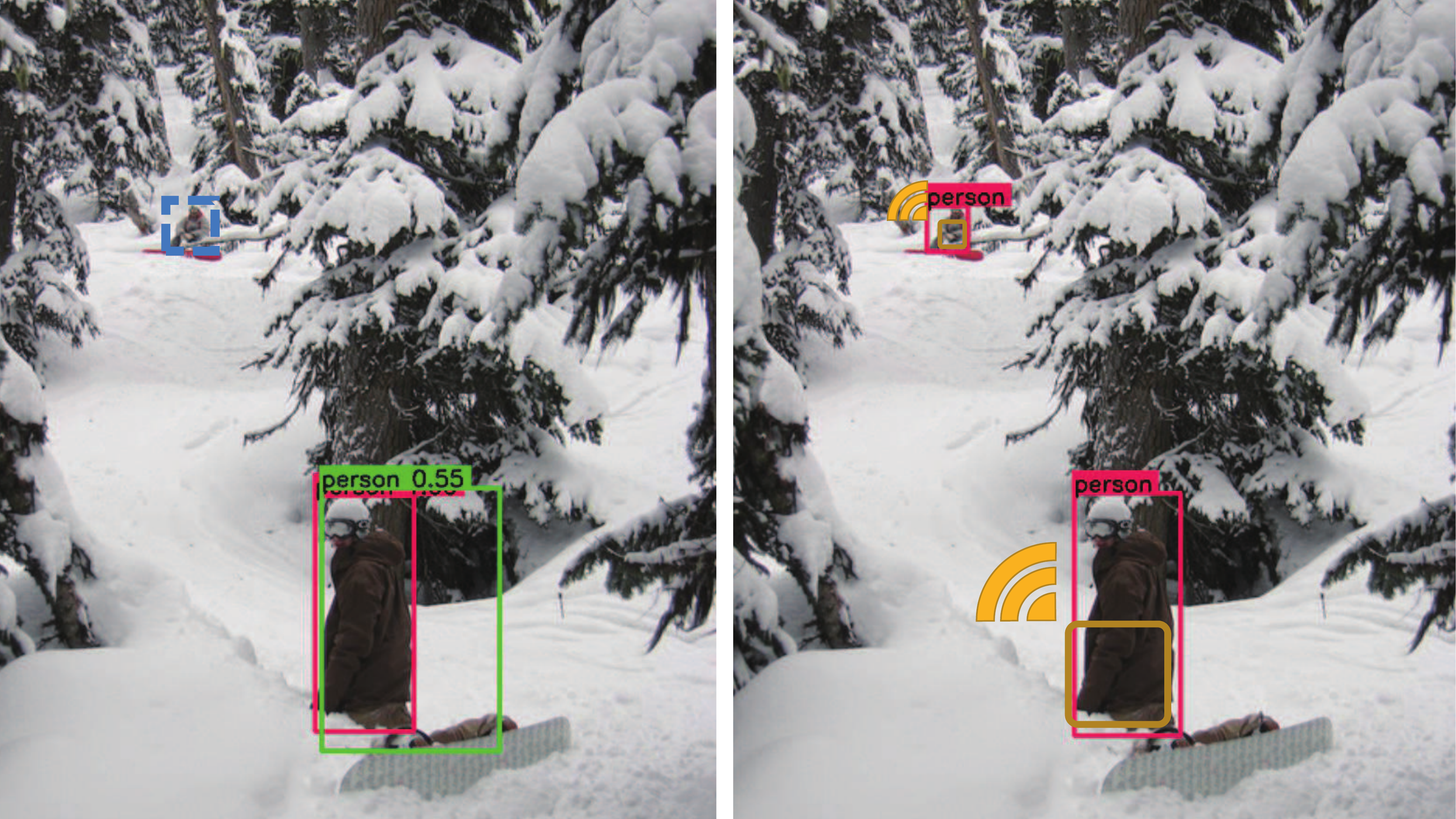}\\
  \caption{The false positives (false detections) and false negatives (miss detections) still exist in the state-of-the-art detection methods (left figure). The false positives stand for the bounding boxes that do not correctly cover a person or cover a person which has been already detected, \eg, the green bounding boxes, while the false negatives represent the miss detections for the persons, \eg, the blue bounding boxes. With the aid of radio information, \eg, the orange regions, the proposed method can well alleviate the false positives and false negatives (right figure).}\label{fig1}
\end{figure}

However, problems still exist in the existing detection methods. As shown in Figure \ref{fig1}, there are mainly two problems: false positives and false negatives. The false positives stand for the bounding boxes that do not correctly cover an object or cover an object which has been already detected, \eg, the green bounding boxes, while the false negatives represent the miss detections for the objects, \eg, the blue bounding boxes. Note that most of the state-of-the-art detection methods use the confidence score to filter the detections, \ie, if the confidence score is larger than a pre-defined threshold, the detection will be displayed in the image. Thus, the false positives are those ``incorrect" bounding boxes with confidence scores larger than the pre-defined threshold, while the false negatives are those ``correct" bounding boxes but with confidence scores smaller than the pre-defined threshold.

On the other hand, with the development of wireless communications and internet of things, radio signals are nowadays pervasive in our daily life. For example, people always carry smart phones, and more and more objects are attached with RFID to track their statuses \cite{RFID1,RFID2,RFID3}. Such radio signals can provide us a unique identifier and localization information for each person. The identifier can be the MAC address or the RFID information, while the localization can be estimated through the angle of arrival (AoA) and time of flight (ToF) information extracted from the channel state information (CSI) of the radio signals. The question now is how to utilize the identifier and localization information to improve the detection performance?

In this paper, we propose a radio-assisted human detection framework by incorporating the identifier and localization information obtained from the radio signals. We first align the localization (\ie, the AoA and ToF) information with the image according to the camera's effective focal length (EFL) and field of vision (FOV), and obtain an initial estimation of each person.
For both the anchor-based one-stage detectors and two-stage detectors, we generate different ways to revise the detection's confidence score based on the localization results.
For the two-stage detectors, we further propose to replace the region proposal network (RPN) with multi-scale anchors generated from the initial estimation as proposals.

Moreover, we utilize the number of identifiers as a constraint for the number of persons to be detected. Based on the specific constraint, we propose an improved non-maximum suppression (NMS) to guarantee that the detections are born from different localizations, and the number of the detections will not exceed the number of the localizations. It is worth pointing out that, with the proposed NMS, there is no need to use the confidence threshold to process the visible detections. Thus, the proposed framework can greatly alleviate the problems of false positives and false negatives in final detections.

We evaluate the proposed method on both simulative datasets and real-world scenarios. The simulative datasets are constructed based on the COCO and Caltech pedestrian datasets. Specifically, we reshape the ground-truth bounding boxes into squares and randomly shift the positions and sizes to simulate the initial estimation from the radio localizations. The experimental results show that, compared with the state-of-the-art detection methods, the proposed method can achieve better detection results in terms of both the mAP metric and the miss rate (MR) versus false positives per image (FPPI) metric. The experimental results on the real-world scenarios also demonstrate that the proposed method can greatly alleviate the problems of the false positives and false negatives, \ie, the false and miss detections.

\section{Related Work}

\textbf{Object detection:}
The state-of-the-art object detectors can be classified as either two-stage detectors \cite{MaskRCNN,RCNN,FasterRCNN,GridRCNN,LibraRCNN} or one-stage detectors \cite{YOLO9000,SSD,Densebox,Extremenet,Cornernet,FCOS}. For two-stage detectors, the first stage is to filter out the background anchors to generate a sparse set of object proposals, while the second stage is to classify the proposals as well as refining the bounding boxes to obtain the final detections. Early two-stage detectors such as R-CNN \cite{RCNN} select regions of interest with traditional algorithms to construct proposals, while the latest two-stage detectors are equipped with RPN for more efficient and accurate proposals \cite{MaskRCNN,FasterRCNN}. Although the development of two-stage detectors may differ in multi-scale features \cite{LibraRCNN,FPN,TMM1}, regression layers \cite{GridRCNN} and balance training \cite{LibraRCNN,OHEM}, their region proposals are all similar, which can be replaced by the initial estimations from the radio signals.

Some one-stage detectors are anchor-based methods \cite{YOLO9000,SSD}, whose idea is to classify and refine each cell's multi-scale anchors and output the final detections with NMS. For example, YOLO\cite{YOLO9000} divides the image into multiple cells and directly refines the pre-defined multi-scale anchors located at each cell as the detection bounding boxes. For the one-stage detector, the anchor stands for the proposal boxes.
Recently, anchor-free detectors have raised more and more attention, some of which rely on keypoint detection to output the bounding boxes \cite{Extremenet,Cornernet}, while others mainly tackle the problem by dense prediction to predict the center of each object \cite{Densebox,FCOS}.

There have been works focusing on object detection tasks with extra information. For example, \cite{Context1} and \cite{Context2} use extra context from the image to improve the detection performance. The authors in \cite{RGBD} utilize the depth image as extra information while the authors in \cite{RGBT} propose to utilize the extra thermal data. Generally, the performance can be improved when the extra information is carefully utilized.

Because the state-of-the-art detectors use confidence score (classification score) to estimate the detection, for the average precision (AP) metric, if the confidence scores of the false positives are lower than those of the true positives, the detection is recognized as a good result.
However, in such a case, false positives with enough confidence displayed in the image will still bother the users with their observation.
By simply rising the confidence score threshold to process the result, some correct detections with lower score will be removed instead.
Therefore a detector performs well in COCO dataset with the mAP evaluation metric may not provide clear visual result.

\textbf{Radio localization:}
The radio localization methods can be classified as received signal strength indication (RSSI) based approaches \cite{RSSI1,RSSI2,RSSI3}, AoA based approaches \cite{AoA1,AoA2,AoA3}, and ToF based approaches \cite{ToF1,ToF2,ToF3}. All these approaches measure the RSSI, AoA, or ToF from the target at multiple antennas and localize the target through triangulation.
The RSSI based approaches measure the RSSI from the target at multiple access points and locate the target by combining the RSSI via triangulation with a propagation model.
With RSSI data of multiple access point, the system can work out the distance of the object to each antenna and finally achieve the coordinate of the object.
The ToF localization is similar as the RSSI localization, which also uses the distance of the object to each antenna to get the object's destination.
The AoA based approaches work out the AoA of the direct path of each receiver to the target.
Similar as the RSSI based methods, the AoA based approaches require multiple estimations at different access points and use the triangulation to localize.

The joint AoA-ToF estimation is considered in \cite{bua,bub,buc,BreathTrack,RoArray}, where the localization of the target can be estimated with the AoA and ToF from one single antenna.
If the environment is equipped with a single access point, we can provide the accurate localization with the joint AoA-ToF estimation.
The position of the object can be worked out with the angle and the distance of the object from the antenna receiver.


\textbf{Radio-video fusion:}
Several literatures have attempted to combine radio signals with video data for tracking/localization tasks \cite{Deep-R-V,People-tracking,RGB-W,RFPose,RFPose2,BU1,BU2,BU3}.
Ishihara \etal in \cite{Deep-R-V} proposed a network for BLE signals and integrated it with the PoseNet.
\cite{People-tracking} focused on minimizing the root mean squared error of the predicted tracking.
Their method shows better accuracy for the camera's indoor position and rotation.
In \cite{RGB-W}, the authors fused the vision and wireless modalities and improved the localization and tracking of individuals.
With the received signal strength (RSS) from individual's mobiles, individual's tracking and localization can be accomplished with the RGB view and a corresponded wireless ring image. Zhao et al. implemented an easy-to-deploy system with multi-modal data for indoor localization~\cite{BU1}, which achieved 92-percentile error within 0.2m for indoor targets. Wang et al. performed localization with images and RFID tags and achieved 6.23cm localization error~\cite{BU2}, while Bai et al. proposed visible light communication-assisted indoor localization that achieved less than 3cm positioning error\cite{BU3}.
Zhao \etal in \cite{RFPose} trained the RF encoder and decoder networks for human pose estimation with the image-based keypoint method as supervision, while Li \etal in \cite{RFPose2} proposed a neural network to predict the human's pose with the input of both RGB image and radio heatmap.


Different from the aforementioned radio-video fusion work, our work focuses on utilizing the identifier and localization information extracted from the radio signals to improve the performance of the state-of-the-art object detection methods.

\section{Our methods}
\subsection{Radio localization and imaging}
\begin{figure}[htbp]
	\centering
	\includegraphics[width=1\linewidth]{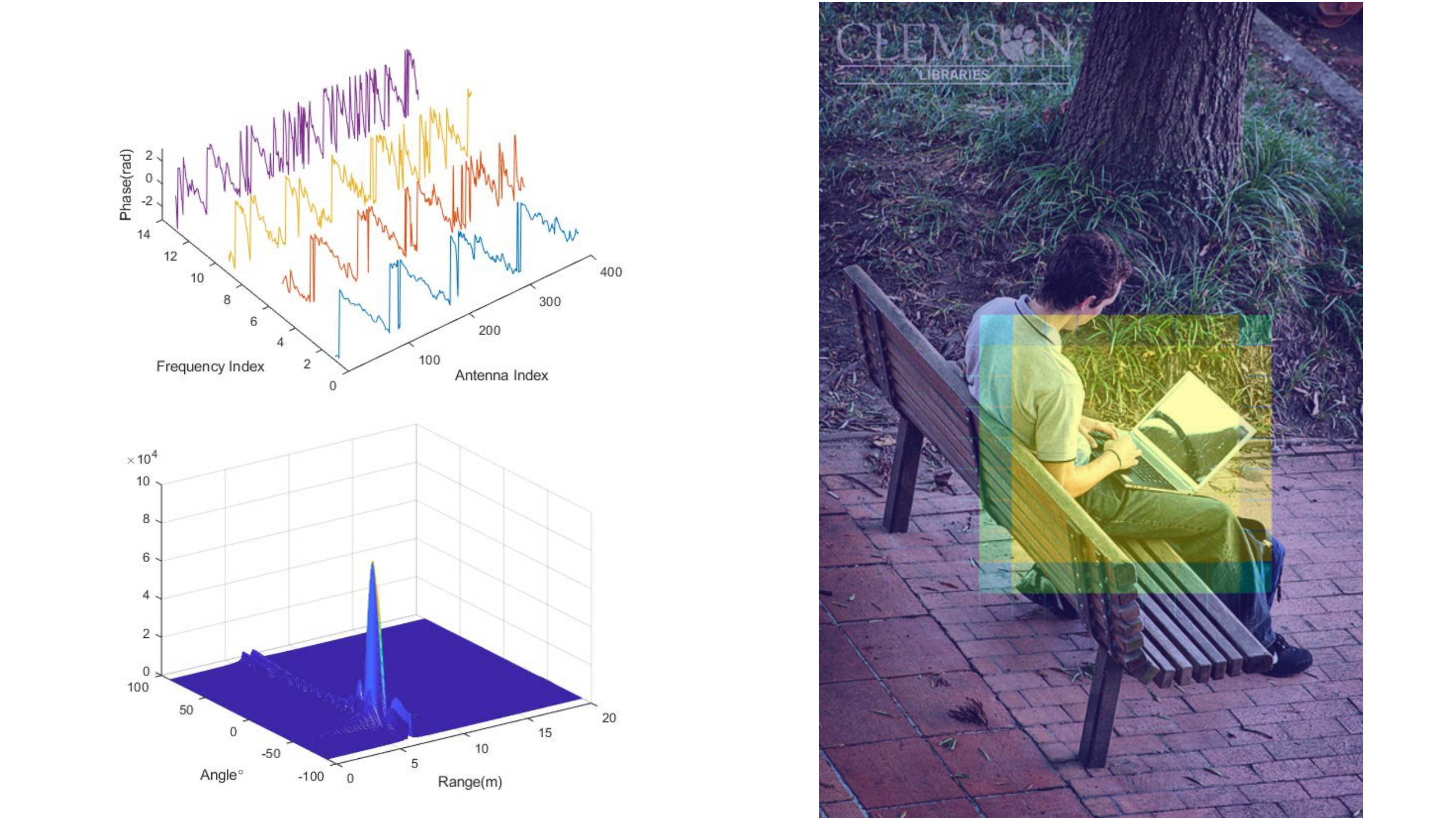}\\
	\caption{Joint AoA-ToF estimation from radio signals. 
	\textbf{Left:} The raw CSI is sampled on different frequencies and antennas shown in the figure above. It is further transformed into AoA-ToF domain as shown in the figure below. The peak corresponds to the device location with highest confidence. 
	\textbf{Right:} An initial estimated region with the radio information. }
	\label{aoatof}
\end{figure}

\textbf{Joint AoA-ToF estimation.}
We assume that each person to be detected is equipped with an RF component that can emit radio signals. This assumption is reasonable given the development of wireless communications, \eg, people nowadays always carry smart phone. The receiver is equipped with a vertical and horizontal antenna array, which is capable of estimating vertical and horizontal AoA of the person. The signal from a person with a specific AoA-ToF can be expressed as
\begin{equation}\label{aoatofeqn}
 P(\theta, \tau) = \sum_{m=0}^{M} \sum_{k=0}^{K} s_{m,k} e^{j2\pi f_k \frac{mdcos \theta}{c}} e^{j2\pi k\Delta f \tau},
\end{equation}
where $\theta$ and $\tau$ denote the AoA and ToF, respectively, $m$ and $k$ are the indexes of antenna and frequency, $d$ is the inter-element space of antenna array, $c$ is the speed of light and $\Delta f$ is the frequency interval.
Then, the AoA-ToF can be estimated by
\begin{equation}
	(\hat{\theta}, \hat{\tau}) = \mathop{\arg\max}_{\theta, \tau} |P(\theta, \tau)|, 
\end{equation}
where $(\hat{\theta}, \hat{\tau})$ denotes the estimated AoA-ToF. In other words, we could estimate the AoA-ToF by pick the peak with the highest amplitude shown in Figure \ref{aoatof}.


\begin{figure}[htbp]
  \centering
  \includegraphics[width=1\linewidth]{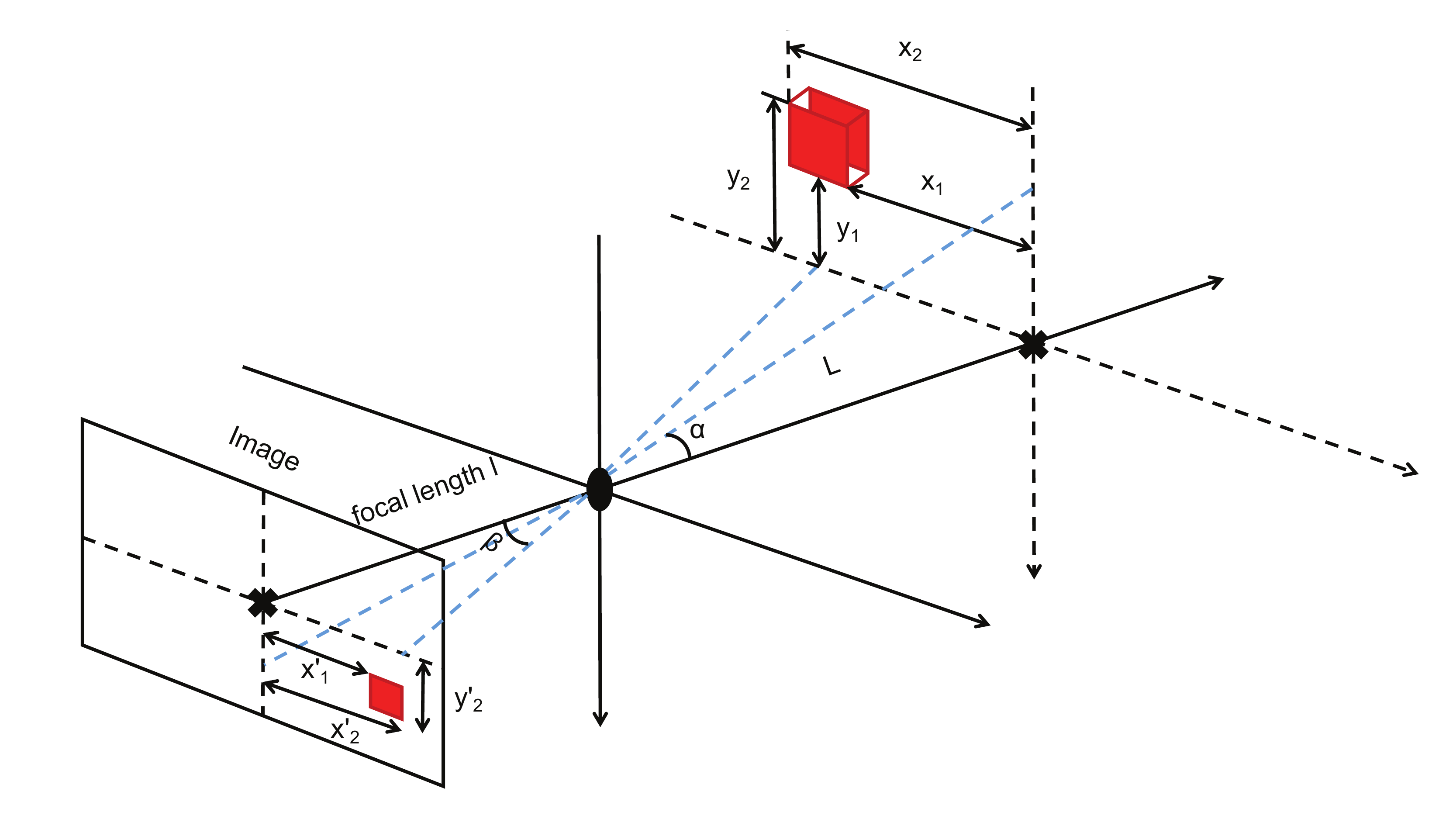}\\
  \caption{Radio imaging to the image plane with the estimated AoA and ToF.}\label{imaging}
\end{figure}

\textbf{Radio imaging.}
With the estimated AoA-ToF, the 3D locations of the person to the camera can be obtained with the camera imaging model. 
Specifically, with the estimated ToF, we first derive the distance of the person to the camera, as shown in Figure \ref{imaging}. Then, we calculate the point-to-plane distance $L$ with the horizontal and vertical AoA, $\alpha$ and $\beta$. With the ratio of $L$ to the camera's focal length $l$, the sizes of the estimated regions in the image can be calculated. The coordinate of the person can then be known with the horizontal angle, the vertical angle and the focal length by using the tangent. In this way, we can obtain the initial estimate of the area occupied by a person.

\begin{figure*}
  \centering
  \includegraphics[width=\linewidth]{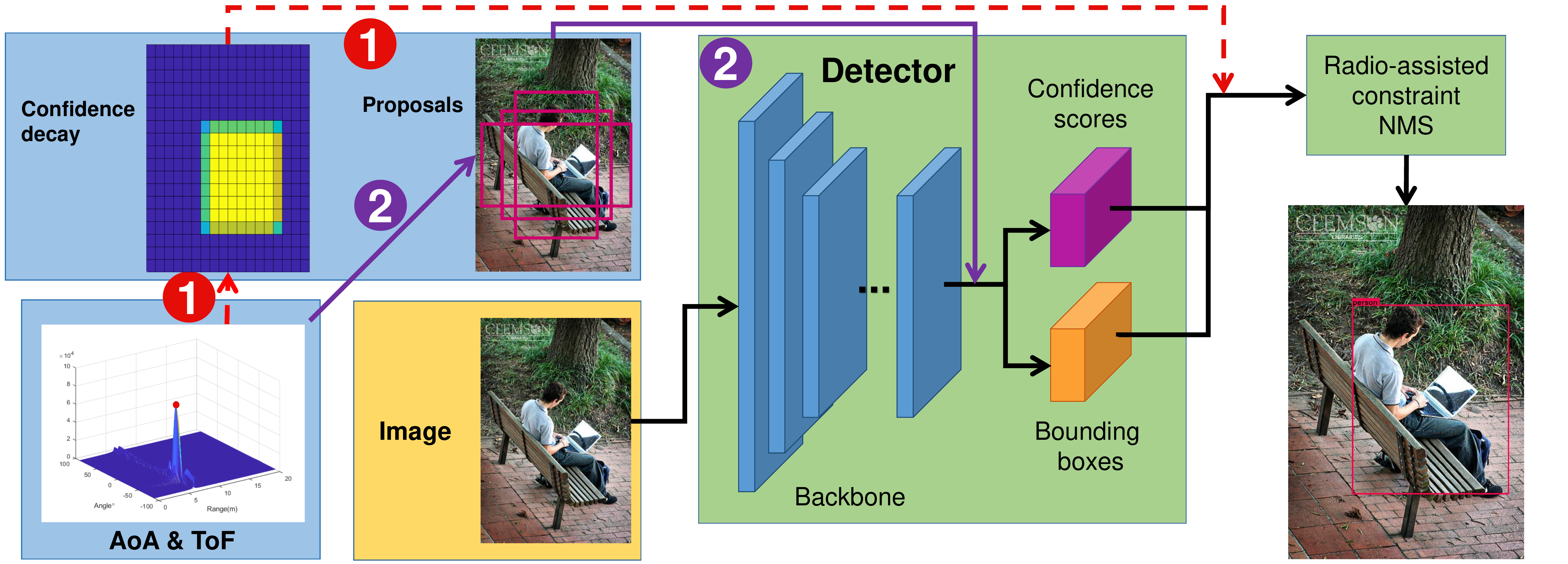}\\
  \caption{Pipeline of the radio-assisted human detection. \textbf{Method 1:}  The pipeline of the detectors with radio-assisted confidence revision. \textbf{Method 2:} The pipeline of the two-stage detector with radio-assisted proposals.}\label{fig3}
\end{figure*}

\subsection{Radio localization aware detectors}
While the existing detection methods have achieved state-of-the-art detection performance, the false positives and false negatives problems still commonly exist in the final detection results. In this subsection, we discuss how to incorporate the radio localization information into the detectors to alleviate the false positives and false negatives, the details of our proposed radio localization guided methods will be introduced, respectively. The pipeline of the whole proposed radio-assisted human detection is shown in Figure \ref{fig3}.

\textbf{Method 1: Radio confidence revision.}
We utilize the localization information from radio signals to revise the confidence scores of the detection results before NMS. In general, if the bounding box correctly covers the target, its confidence score should be larger than those of the boxes which only partly include the target.
Thus, we introduce a decay factor, $\gamma\in[0,1]$, for each detection bounding box.
For anchor-based one-stage detectors, given a radio-assisted region in an image, $\gamma$ is defined as the normalized intersection between the radio region and the detection's corresponding divided cell in the backbone over the cell, \ie,
\begin{eqnarray}\label{decay1}
\gamma = \frac{area(localization)\bigcap area(cell)}{area(cell)}.
\end{eqnarray}

For two-stage detectors, we adjust each detection's confidence score using the intersection between the bounding box and the localization area over the localization area, \ie,
\begin{eqnarray}\label{decay2}
\gamma = \frac{area(bbox)\bigcap area(localization)}{area(localization)}.
\end{eqnarray}

Then, the new confidence score is obtained as follows
\begin{eqnarray}
S_{new} = (1-\lambda+\lambda*\gamma)*S,
\end{eqnarray}
where $\lambda$ is a pre-defined constant defined by the expected accuracy of the radio localization, \ie, a larger $\lambda$ means a higher expected accuracy. The case where $\lambda=0$ reflects that the method is only determined by the traditional detection framework, while the case where $\lambda=1$ means the revised score is processed by directly multiplying with the decay factor.

\textbf{Method 2: Radio region proposals for two-stage detectors.}
For the two-stage detectors, we can directly replace the RPN with the radio region proposals. The RPN in two-stage detectors is in charge of selecting and refining proposals from multi-scale anchors to filter out the proposals of the background. For our detectors, the radio localization provides the multi-scale anchors as region proposals whose centers and sizes are determined by the radio imaging discussed in the previous subsection. If we only focus on the single-class person detection, the anchor shape and ratio can be designed in advance based on the people's potential postures and general size. With the proposed method, by refining the anchors only once, the detector is capable of outputting accurate detections. Note that by replacing the RPN with the radio region proposals, our method is capable of speeding up the two-stage detectors.
\begin{figure}
  \centering
  \includegraphics[width=\linewidth]{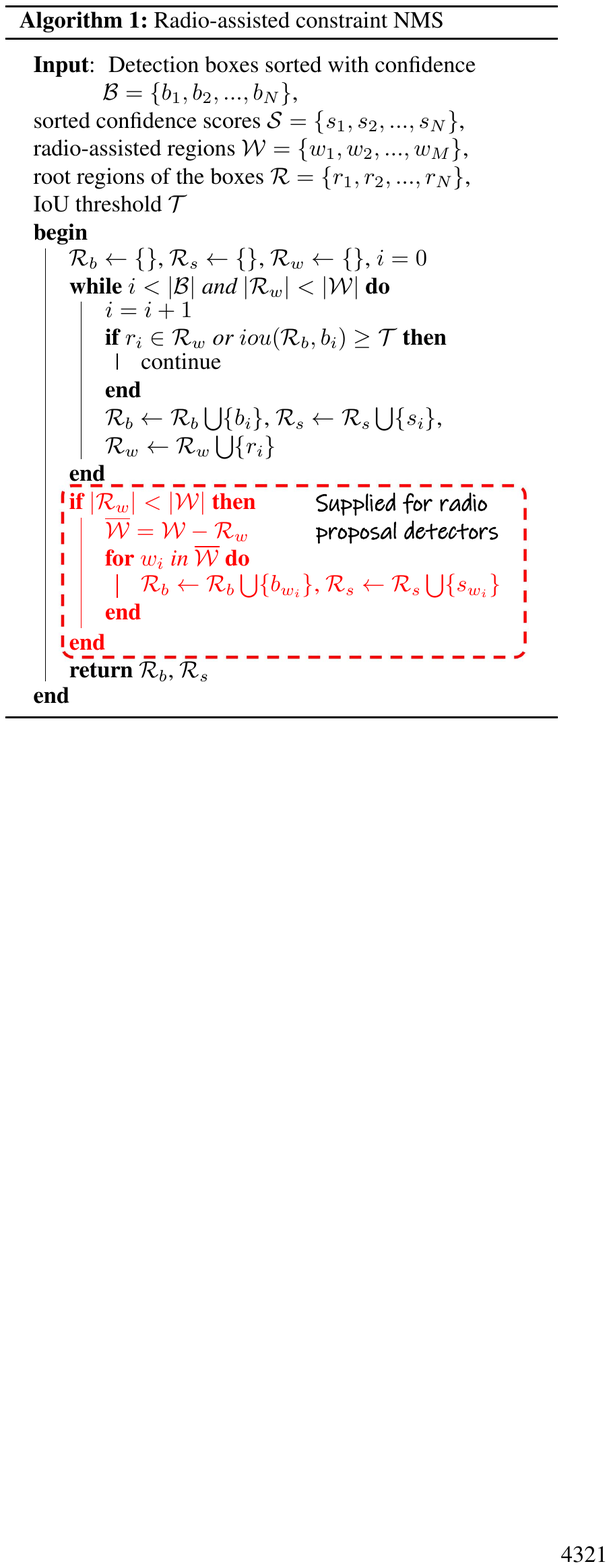}\\
  \caption{NMS with the one-on-one constraint of the radio-assisted region and the detection bounding box. The procedure marked red is only feasible for detectors with the radio region proposal input. }\label{Algorithm}
\end{figure}

\subsection{NMS with radio region constraint}
An NMS method with the number constraint has been proposed in \cite{C-WSL}, where the constraint is not from the radio localization. In \cite{C-WSL}, the authors choose the final detections by maximizing the sum of their scores with the constraint of the scores' count and the Intersection over Union (IoU) of the detections. 
IoU specifies the amount of overlap between the predicted and ground truth bounding box, which is an important metric to evaluate the performance of object detection.
In this paper, we consider the constraint that each bounding box should be born from different radio localizations. The radio localization constraint could be recognized as a reinforced constraint of the number of persons because each localization finally will be matched with at most one bounding box. The details of the proposed radio region constraint NMS algorithm is illustrated in Figure \ref{Algorithm}.

In the proposed NMS algorithm, the first loop is a similar process as the common NMS. The main improvement of the first loop is that one radio localization will at most output one bounding box. The condition $iou(\mathcal{R}_b,b_i)\geq \mathcal{T}$ represents that the IoU of $b_i$ and any bounding box in $\mathcal{R}_b$ is not smaller than the threshold $\mathcal{T}$. The condition $r_i\in\mathcal{R}_w $ refers that the radio region $r_i$ of the current detection $b_i$ has already existed in the result set, which is set to satisfy the constraint that one radio localization at most matches one detection bounding box. With the first procedure, it can be guaranteed that the neighbourhood boxes all lie out of the pre-defined overlap and different bounding boxes are born from different radio localizations.

For the two-stage detectors with the radio region proposal input, it is easy to know each detection comes from which proposal and belongs to which radio region. For the one-stage detectors with radio confidence mask, it is difficult to know the responsible radio region because the detections are the end-to-end results and one cell/detection may be covered by multiple radio regions. In such a case, we formulate the relation by the IoU of each single radio region and the detection box.

The second loop in Figure \ref{Algorithm} is enabled if the number of the detections is smaller than the number of the radio localizations. This may happen when some correct detections are suppressed by unreasonable IoU threshold. With the second loop, the algorithm will output the detection generated from a pre-defined anchor of the missing radio region. This loop is implemented only for the two-stage detectors with the radio proposal input. For the one-stage detectors, it cannot be guaranteed that every region of radio localization is allocated at least one detection with the IoU relation. In such a case, we skip the second process and the proposed NMS can only ensure that the number of the final detections will not exceed the number of the radio localizations.

\section{Experiments}

Extensive experiments on simulative datasets and real-world scenarios are conducted to verify the effectiveness of our method by comparing with the state-of-the-art detection methods.
For experiments on detection datasets, we simulate the wireless localization region from the ground truth and conduct detections.
Firstly, we study the influence of the localization deviation (AoA and ToF) on the detection precision.
We then compare the performance of existing detectors with our radio localization aware detectors on these datasets. Note that as shown in section 3.2, we have proposed two radio localization aware detectors: one is radio confidence revision, denoted as ``\textbf{Proposed Method 1}"; the other is radio region proposal, denoted as ``\textbf{Proposed Method 2}". Specifically, for two-stage detectors equipped with radio region proposal, we choose the region in the image mapped from the radio localization and design three scales of the anchor to match the human potential postures.
For the radio confidence revision, based on the localization region, we adjust each detection's confidence score as (\ref{decay1}) and (\ref{decay2}).
Finally for real data, the wireless localization results together with the detection results in a real-word scenario are presented and the detections of the existing detectors are also illustrated for comparison.


\subsection{Experimental Setup}
\textbf{Datasets:}
The simulative datasets are constructed based on COCO \cite{COCO} and Caltech pedestrians \cite{Caltech} datasets. For COCO, we conduct detection tasks for 80 categories and  person category and evaluate the detectors on the \emph{val2017} set. For Caltech pedestrian, we implement human detections on every 30 frame in the video sequences in validation \emph{set06} to \emph{set10}.

In practice, the AoA and ToF estimated from the radio localization system may deviate from the ground truth due to the multipath interference in the environment.
Since we do not have the prior knowledge about the shape of the person, we first reshape the ground-truth bounding box into a square region whose edge length $L$ equals $\min(H,W)$. Then, to imitate the ToF estimation errors in radio localization, we manufacture independent Gaussian distributed noise $\zeta$ to the edge length $L$ of each square region as below
\begin{eqnarray}\label{sigma}
L'=L\times\zeta,&&\zeta\sim \mathcal{N}(1,\sigma),
\end{eqnarray}
where $\sigma$ is a pre-defined standard deviation, $\mathcal{N}$ denotes the Gaussian distribution.
Finally, to imitate the AoA estimation errors, we give the center of each square
region a random shift related to its edge length $L'$ as
\begin{eqnarray}\label{k}
x'=x+\xi_1, &&y'=y+\xi_2,\nonumber\\
\xi_1\sim \mathcal{N}(0,k_1L'),&&\xi_2\sim \mathcal{N}(0,k_2L'),
\end{eqnarray}
where $\xi_1$ and $\xi_2$ are the random shifts along the $x$ and $y$ direction, respectively, $k_1$ and $k_2$ are pre-defined standard deviations.

\textbf{Metrics:} We adopt two widely used metrics for detection performance: the mean average precision (mAP) defined in COCO \cite{COCO} and the miss rate (MR) versus false positives per image (FPPI) curve defined in Caltech pedestrian \cite{Caltech}. The mAP evaluates the mean value of the AP with the requirement of the IoU interval [.5:.05:.95] between the output and ground-truth bounding boxes. The MR vs FPPI curve draws the MR curve under different FPPI and finally obtains the average MR to evaluate the detectors. The runtime is also evaluated to demonstrate that our method could improve not only the detection performance but also the detection efficiency.

\begin{figure}
  \centering
  \subfigure[Effect of ToF estimation errors (person category).]{
  \includegraphics[width=1\linewidth]{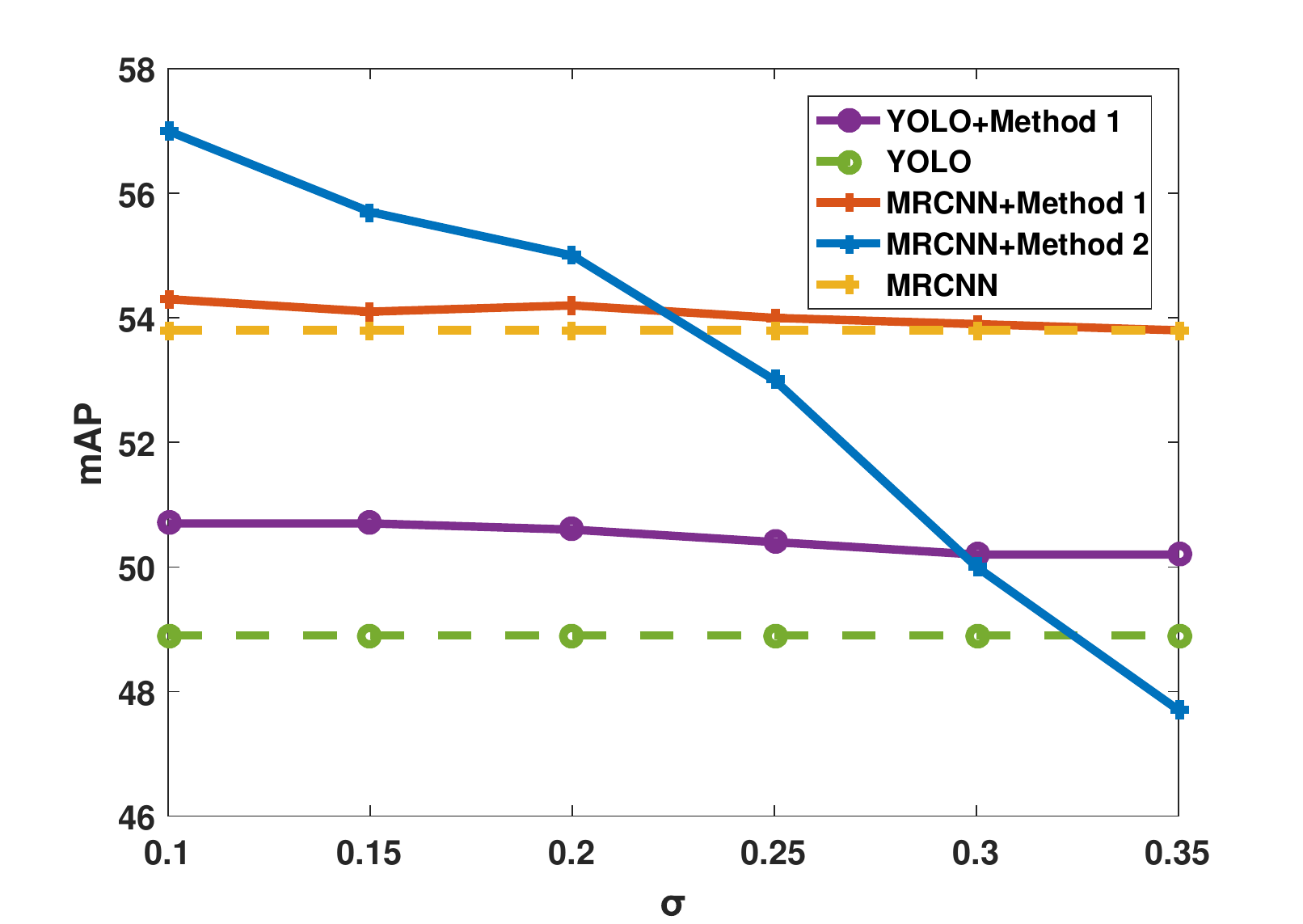}
  \label{NoiseToF}}
  \subfigure[Effect of AoA estimation errors (person category).]{
  \includegraphics[width=1\linewidth]{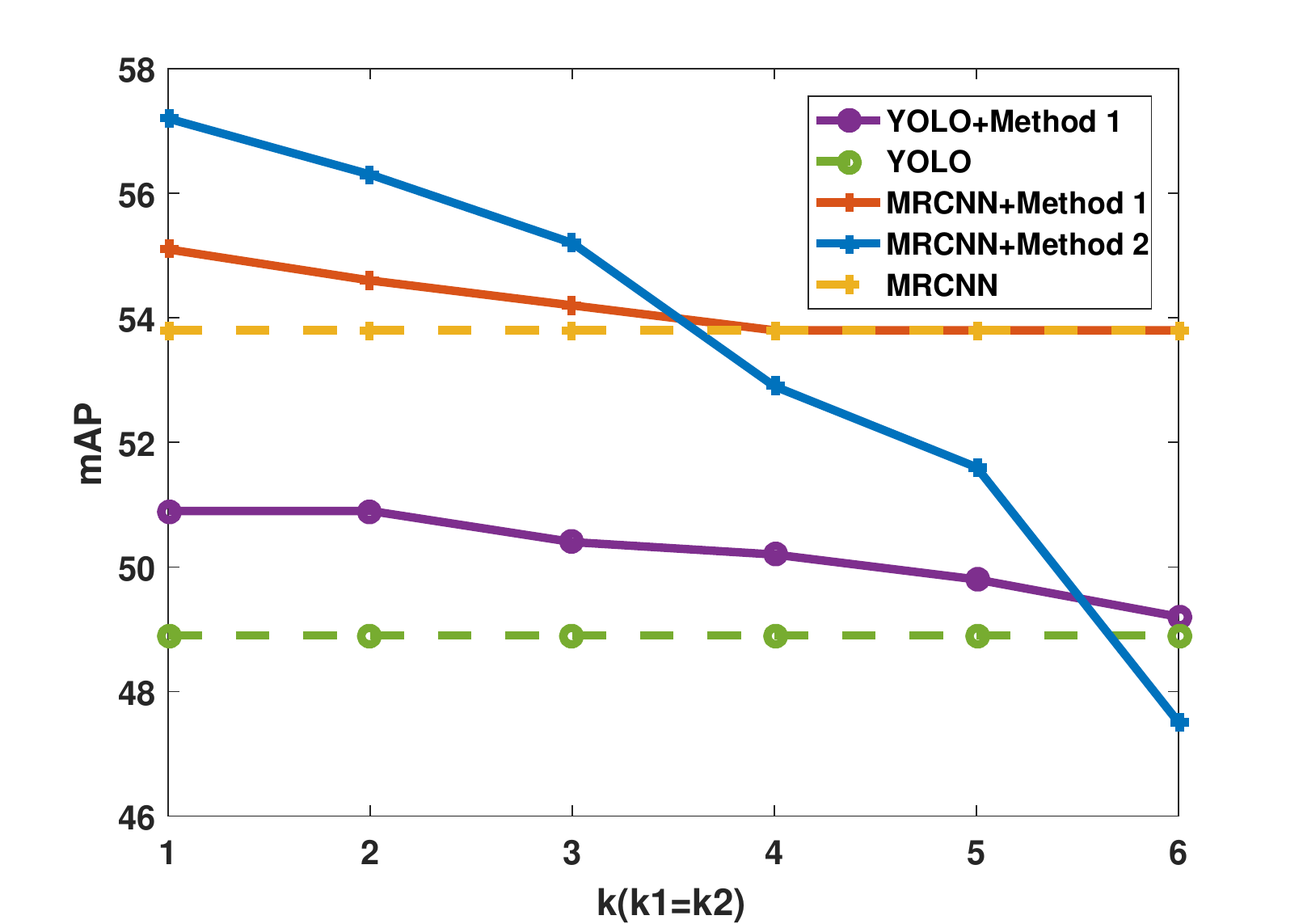}
  \label{NoiseAoA}}\caption{Effect of radio localization error.}\label{Noise}
\end{figure}

\textbf{Implementation details:}
We apply our method on various two-stage detectors including Mask R-CNN, Libra R-CNN with the backbone ResNet-101 and Grid R-CNN with the ResNext-101-32x4d backbone. We adopt the model weights in \cite{mmdetection}. The confidence score threshold and the IoU threshold of the NMS in Mask R-CNN and Libra R-CNN are 0.05 and 0.5, respectively. In Grid R-CNN, the NMS's confidence threshold is 0.03 and the IoU threshold is 0.3 based on \cite{GridRCNN+}. We also apply our method on the anchor-based one-stage detector YOLOv3 on Keras with the model weight in \cite{YOLOv3}. The backbone is Darknet-53 in \cite{YOLOv3} with the $608\times608$ input image size and the NMS's confidence threshold and the IoU threshold being 0.05 and 0.5 respectively. We perform the experiments on a single Geforce GTX 1080Ti GPU to measure the runtime.

\subsection{Effect of radio localization error}
We first analyze the effect of radio localization error on the mAP when the simulative COCO dataset is used.
The influences of scale to the region size and shift to the region center are illustrated in Figure \ref{NoiseToF} and Figure \ref{NoiseAoA} respectively.
When conducting the analysis of $\sigma$ or $k(k_1=k_2)$ in Eqn. (\ref{sigma}) or (\ref{k}), the other parameter is set as $k=0.1$ or $\sigma=0.2$.
By comparing Figure \ref{NoiseToF} and \ref{NoiseAoA}, we observe that the AoA error has a more significant influence on the mAP than the ToF error.
Therefore, it's more important to provide a precise coordinate of each person's center by accurate AoA estimation.
From the figure, we can also see that with higher accuracy, the radio-assisted radio proposal in two-stage detectors performs better than the confidence adjustment.
While on the other hand, the radio-assisted confidence revision is more robust than the radio-assisted region proposal at the lower localization accuracy.
This phenomenon may be due to the fact that two-stage detectors crop features in the proposals to the classification and regression layers, and thus the region proposal requires higher localization precision.

\begin{table*}[]
\caption{Ablation studies of the radio-assisted detectors on COCO \emph{val-2017} for \emph{person} category.}\label{Table1}
\centering
\begin{tabular}{llccccccccc}
\hline
Methods & radio localization &   AP & AP$_{50}$ & AP$_{75}$ & AP$_S$ & AP$_M$ & AP$_L$\\ \hline
Mask R-CNN        &                                    &  53.8  &    83.3    &   58.4     &   36.3    &  \textbf{61.8}     &   \textbf{70.1}    \\
&         proposed method 1                          &  54.2   &   86.0      &  58.8      &  39.7     &  61.7     &  67.8   \\
        &         proposed method 2                           &  \textbf{55.8}  &    \textbf{87.0}    &   \textbf{61.2 }    &   \textbf{41.1}    &  61.3     &   68.6    \\
         \hline
Grid R-CNN        &                                    &  56.5  &    82.8    &   60.7     &   38.4    &  63.3     &   \textbf{74.4}    \\
&         proposed method 1                          &  \textbf{57.7}   &   \textbf{85.4}      &  \textbf{62.2}      &  \textbf{42.2}     &  \textbf{64.0}     &  73.7   \\
        &         proposed method 2                           &  56.5  &    83.0    &   61.0    &    40.7   &   61.8    &    71.2   \\
         \hline
Libra R-CNN        &                                   &  54.7  &    83.3     &  59.6     &   36.3    &   62.0    &   \textbf{71.6}    \\
&         proposed method 1                         &  \textbf{55.3}   &   \textbf{86.4}      &  60.1      &  \textbf{40.3}     &  \textbf{62.2}     &  69.6   \\
        &         proposed method 2                           &  \textbf{55.3}  &    84.5    &   \textbf{61.0}     &   38.9    &   61.5    &   69.6    \\
        \hline
YOLOv3        &                                        &  48.9   &   82.4     &   52.2     &   30.7    &   56.4    &  66.2     \\
        &          proposed method 1                        &  \textbf{50.3}   &   \textbf{84.6}      &  \textbf{53.7}      &  \textbf{33.3}     &  \textbf{58.0}     &  \textbf{66.6}  \\
         \hline
\end{tabular}
\end{table*}

\begin{table*}[]
\caption{Ablation studies of the radio-assisted detectors on COCO \emph{val-2017} for \emph{person} category with detection's count constraint.}\label{Table2}
\centering
\begin{tabular}{lllcccccccc}
\hline
Methods & radio localization & NMS &  AP & AP$_{50}$ & AP$_{75}$ & AP$_S$ & AP$_M$ & AP$_L$\\ \hline
Mask R-CNN        &                 &                    &  49.3  &    74.3    &   54.6     &   30.8    &  57.2     &   68.4    \\
        &         proposed method 1        &                    &  50.1  &    80.6    &   54.0     &   35.4    &   57.4    &   63.7    \\
        &         proposed method 1        &        GRS\cite{C-WSL}            &  49.5  &    77.9    &   54.1     &   35.6    &   57.3    &   61.3    \\
        &         proposed method 1        &         proposed      &     52.0 & 82.7  &    57.1    &   37.7       &   58.9     &   65.5    &  \\
        &         proposed method 2        &                    &  54.5  &    85.1    &   59.3     &   39.8    &  59.8     &   68.1    \\
        &         proposed method 2        &          GRS\cite{C-WSL}         &  52.0  &    80.1    &   56.8     &   36.1    &  58.7     &   64.9    \\
        &         proposed method 2       &         proposed          &   \textbf{58.3}  &    \textbf{93.2}    &   \textbf{63.0}      &   \textbf{45.3}     &    \textbf{63.4}    &  \textbf{70.2}   \\
        \hline
Grid R-CNN        &                 &                    &  52.4  &    75.1    &   56.9     &   33.0    &  59.1     &   72.7    \\
        &         proposed method 1        &                    &  54.8  &    81.2    &   59.3     &   38.5    &   61.2    &   71.0    \\
        &         proposed method 1        &       GRS\cite{C-WSL}             &  51.5  &    75.1    &   56.0     &   36.9    &   58.9    &   65.7    \\
        &         proposed method 1        &         proposed      &     56.3 & 82.7  &    61.1    &   40.6       &   62.3     &   72.5    &   \\
        &         proposed method 2        &                    &  56.3  &    83.1    &   61.1    &    41.4   &   61.3    &    70.8   \\
        &         proposed method 2        &         GRS\cite{C-WSL}          &  53.6  &    79.7    &   57.9    &    38.4   &   59.7    &    66.8   \\
        &         proposed method 2        &         proposed          &   \textbf{59.1}  &    \textbf{88.4}     &     \textbf{63.1}    &   \textbf{43.4}     &   \textbf{64.3}    &   \textbf{74.0}   \\\hline
Libra R-CNN        &                 &                   &  50.6  &    75.3     &  56.6     &   31.3    &   57.8    &   69.9    \\
        &         proposed method 1        &                    &  51.4  &    80.8    &   56.0     &   36.2    &   58.6    &   65.9    \\
        &         proposed method 1        &       GRS\cite{C-WSL}             &  51.1  &    78.4    &   56.2     &   36.7    &   58.6    &   64.0    \\
        &         proposed method 1        &         proposed      &     53.1 & 82.7  &    58.1    &   38.3       &   60.2     &   67.1    &        \\
        &         proposed method 2        &                    &  55.1  &    83.9    &   61.0     &   39.5    &   61.2    &   68.8    \\
        &         proposed method 2        &          GRS\cite{C-WSL}         &  51.6  &    79.5    &   56.4     &   34.8    &   58.7    &   64.9    \\
        &         proposed method 2       &         proposed      &     \textbf{58.9} & \textbf{92.0}  &    \textbf{64.0 }    &   \textbf{44.7 }       &   \textbf{64.3}     &   \textbf{71.9 }    &  \\\hline
YOLOv3        &                 &                  &  46.8   &   77.9     &   50.5     &   28.0    &   54.2    &  65.6     \\
        &         proposed method 1        &                    &  48.9   &   81.4      &  53.1      &  31.6     &  56.3     &  66.0   \\
        &         proposed method 1        &         GRS\cite{C-WSL}         &  47.1   &   79.9      &  49.5      &  30.0     &  54.7     &  63.5   \\
        &         proposed method 1        &         proposed          &    \textbf{50.8}  &  \textbf{86.4}  &   \textbf{54.1}     &   \textbf{34.6}     &   \textbf{58.0}    &   \textbf{66.0}
        \\ \hline
\end{tabular}
\end{table*}

\subsection{Ablation study}
In this subsection, we conduct the ablation experiments to show that both the radio region proposal/confidence revision and our proposed NMS are effective. The parameters of radio localization errors are set as: $\sigma=0.2$, $k_1=k_2=0.1$.

\textbf{COCO mAP evaluation.}
Table \ref{Table1} and Table \ref{Table2} show the ablation results on COCO for person category.
For the human detection as shown in Table \ref{Table1}, we can see that with the help of radio-assisted confidence revision, the mAP performance for person category can be improved about 0.4\%, 1.2\%, 0.6\% and 1.4\% for Mask R-CNN, Grid R-CNN, Libra R-CNN, and YOLOv3, respectively.
With radio localization region proposal, the performance can be raised for about 2\% and 0.6\% for Mask R-CNN and Libra R-CNN, although the Grid R-CNN's performance may be slightly worse.

We also compare the mAP performance in human detection while we constrain the number of the detections by the count of the ground truth instead of using confidence threshold.
In this case we can conduct a fair performance evaluation on our proposed NMS method.
The results are shown in Table \ref{Table2}.
In the results, we find that if the number of the detections is limited, the improvement of the radio-assisted methods is much more obvious.
Furthermore, we can see that equipped with our proposed NMS, the detectors' performance can even increase about 2\% for confidence revision and 3\% for localization's region proposal.
In Table \ref{Table2}, we also make performance comparison with a mentioned NMS in \cite{C-WSL}, which is an NMS for weakly supervised localization with object's count constraint.
And we find that their NMS doesn't perform well in our person category detection tasks for it only optimizes the specific case where detected bounding boxes are loose and
contain two or more object instances.

\begin{figure*}
  \centering
  \subfigure[MR v.s. FPPI on COCO.]{
  \includegraphics[width=0.31\linewidth]{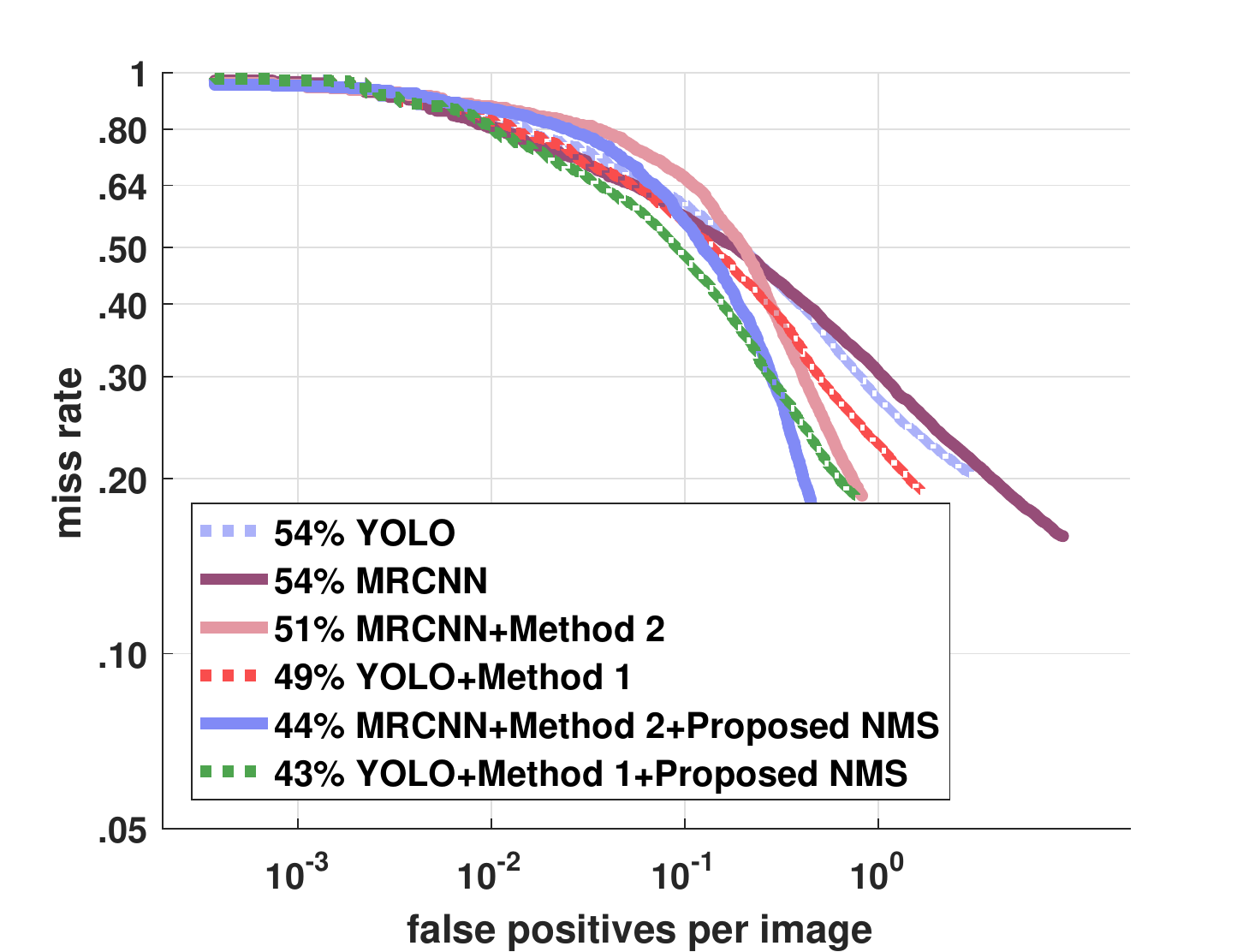}
  \label{MR3}}
  \subfigure[MR v.s. FPPI on Caltech Pedestrian's `reasonable' experiment.]{
  \includegraphics[width=0.31\linewidth]{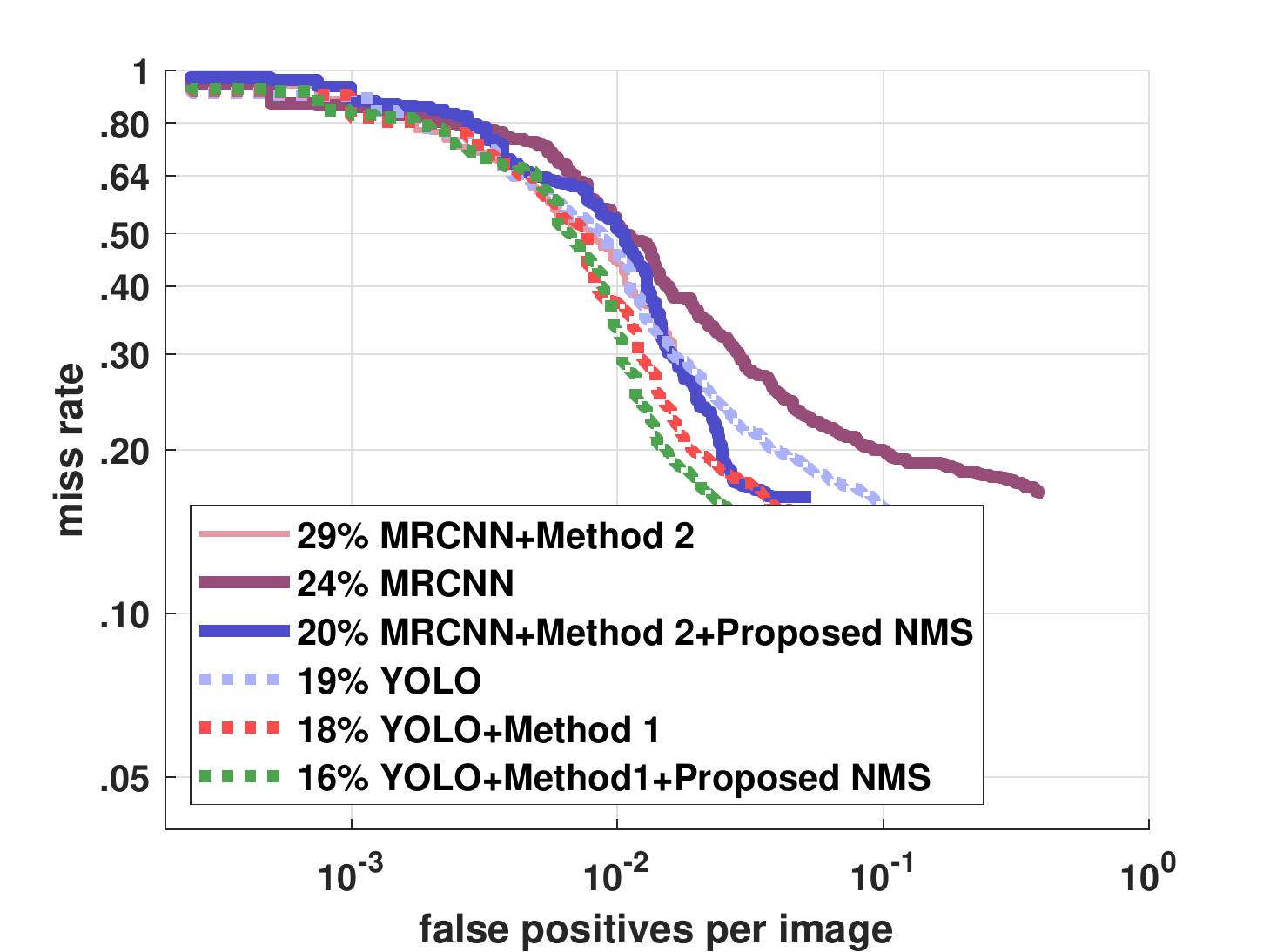}
  \label{MR1}}
  \subfigure[MR v.s. FPPI on Caltech Pedestrian's `all' experiment.]{
  \includegraphics[width=0.31\linewidth]{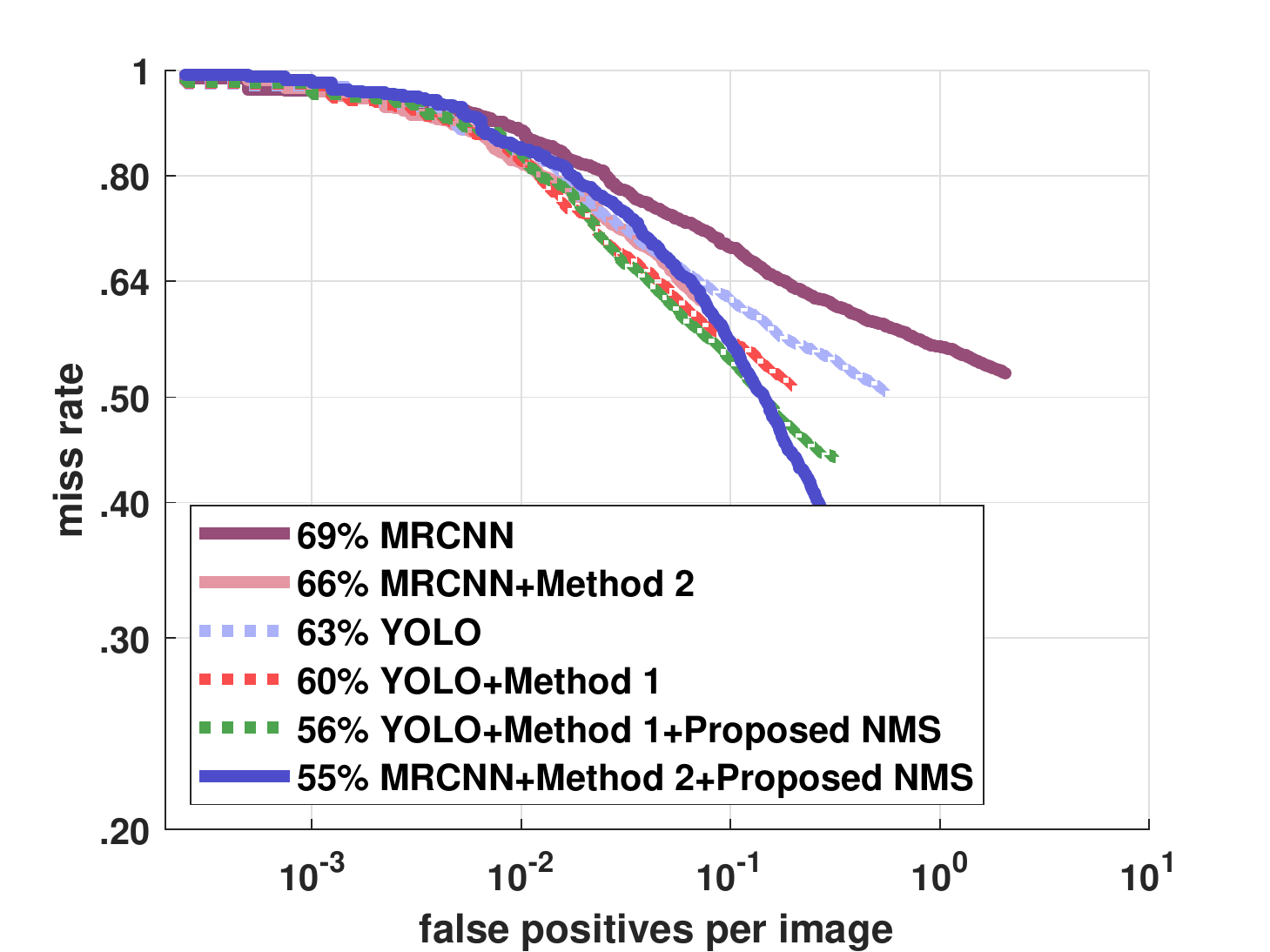}
  \label{MR2}}
\caption{Miss rate versus false positives per image.}\label{MR}
\end{figure*}

\begin{figure*}[]
  \centering
  \includegraphics[width=1.0\linewidth]{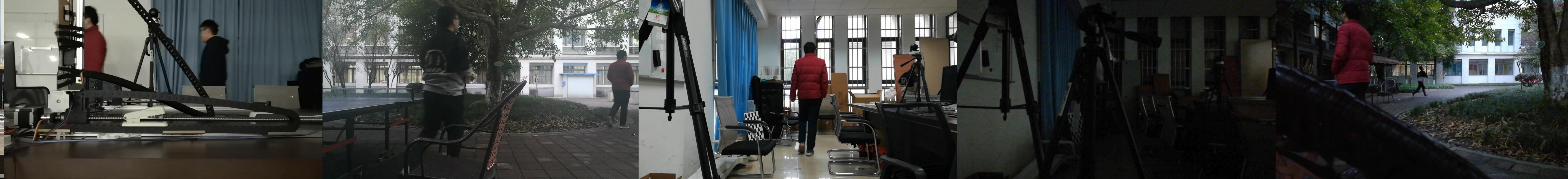}\\
  \caption{Scenarios of the adopted validation dataset.}\label{dataset}
\end{figure*}

\textbf{MR vs FPPI evaluation.}

To give enough penalty on the false positives, we employ the MR v.s. FPPI metric \cite{Caltech}. The MR v.s. FPPI curves of human detection on the COCO dataset are shown in Figure \ref{MR3} to evaluate the ability of the proposed method on suppressing false positives while conducting correct detections, where the numerical percentages in the legend stand for the average miss rates. The results show that the average miss rate and the FPPI at low miss rate are both improved with the proposed method. Compared with the mAP metric in Table \ref{Table2}, the improvement of the average miss rate in Figure \ref{MR3} can be explicitly observed.

We also evaluate the Caltech pedestrian dataset with the `reasonable' experiment and the `all' experiment, respectively. In the `reasonable' experiment, the ground truth only contains people of height more than 60 pixels and occlusion ratio less than 35\%, while in the `all' experiment, detectors should detect people of height more than 20 pixels and occlusion ratio less than 80\%. The results are shown in Figure \ref{MR1} and Figure \ref{MR2}. Similar to the results on the COCO dataset, both the `reasonable' and `all' experiments reflect that the proposed method can efficiently reduce the average miss rate. Furthermore, it can be observed from the figures that with our proposed NMS, the FPPI is significantly decreased at low miss rates, which demonstrates that our proposed NMS can reduce the number of the false positives.

\begin{table*}[]
\centering
\caption{Performance of the radio-assisted detectors on our validation dataset.}\label{OurDataset}
\begin{tabular}{llllll}
\hline
Methods    & Localization & Proposed NMS & FP \& FN per image & {True Detections Ratio(\%)}  \\ \hline
Mask R-CNN &              &              & 1.152              & {46.44}                     \\
           & Method 1     & \checkmark            & \textbf{0.990}              & \textbf{52.18}                     \\
           & Method 2     & \checkmark           & 1.001              & {49.04}                     \\

           \hline
Grid R-CNN &              &              & 1.003              & {48.28}                     \\
           & Method 1     & \checkmark            & \textbf{0.984}              & \textbf{52.42}                     \\
           & Method 2     & \checkmark           & 1.088              & {46.25}                     \\

           \hline
Libra R-CNN &              &              & 1.564              & {42.03}                     \\
           & Method 1     & \checkmark            & \textbf{0.990}              & \textbf{52.82}                     \\
           & Method 2     & \checkmark           & 1.004              & {48.71}                     \\

           \hline
YOLOv3     &              &              & 0.499              & {69.87}                     \\
           & \checkmark           & \checkmark           & \textbf{0.329}              & \textbf{75.03} \\ \hline
\end{tabular}
\end{table*}

\subsection{Experiments in real-world scenarios}

Finally, we conduct experiments in the real-world scenarios to verify the feasibility of our method in practice.

To make detection evaluation with real localization data, we conduct our validation dataset with synchronized image and localization data.
We captured videos in three scenarios with two cameras and the total length of videos is about 15 minutes.
We collected the synchronized localization data with one localization device to work out the coordinates of the people.
The camera we used is the camera of HUAWEI honor8lite with the horizontal FOV 64${}^{\circ}$ and the vertical FOV 52${}^{\circ}$. The sizes of the captured videos are $1280\times720$ pixels and the pixel-measured focal length approximately equals to 3000.
We divided the videos into nearly 1000 frames and some of them are shown in Figure \ref{dataset}.
The environment contains laboratory and open spaces in overcast and evening, and people's activities include walking, jogging, hugging and jumping.
In order to increase the difficulty of detection, we choosed to capture the image data in the scenarios where the brightness is not efficient or some obstacles (\ie chairs and tripods) exist in the image which cover half part of one person at most (\ie head, legs or body).
In one frame there are at most three people and a data frame may also be empty, \ie, it does not include any person.

To collect the localization data, we used the TI board of MMWCAS-RF-EVM to collect the RF signals, and adopted Equation (\ref{aoatofeqn}) to work out the coordinates of the people in the localization space. Finally we calculated the length and the angle of each person to each camera.

To show the main advantage of our method, we use two other evaluation methods in our dataset. Because our methods focus more on the final visual performance in detection tasks, the evaluation methods consider all of the detection bounding boxes without sorting them in descending order by the confidence score.
One evaluation is to calculate the false positives and false negatives per image, the other we called true detection ratio is to calculate the ratio between true positives to the sum of true positives, false positives and false negatives, \ie
\begin{eqnarray}
\frac{true\  positives}{true\  positives+false\  positives+false\  negatives}.
\end{eqnarray}

\begin{figure}[]
  \centering
  \includegraphics[width=1.0\linewidth]{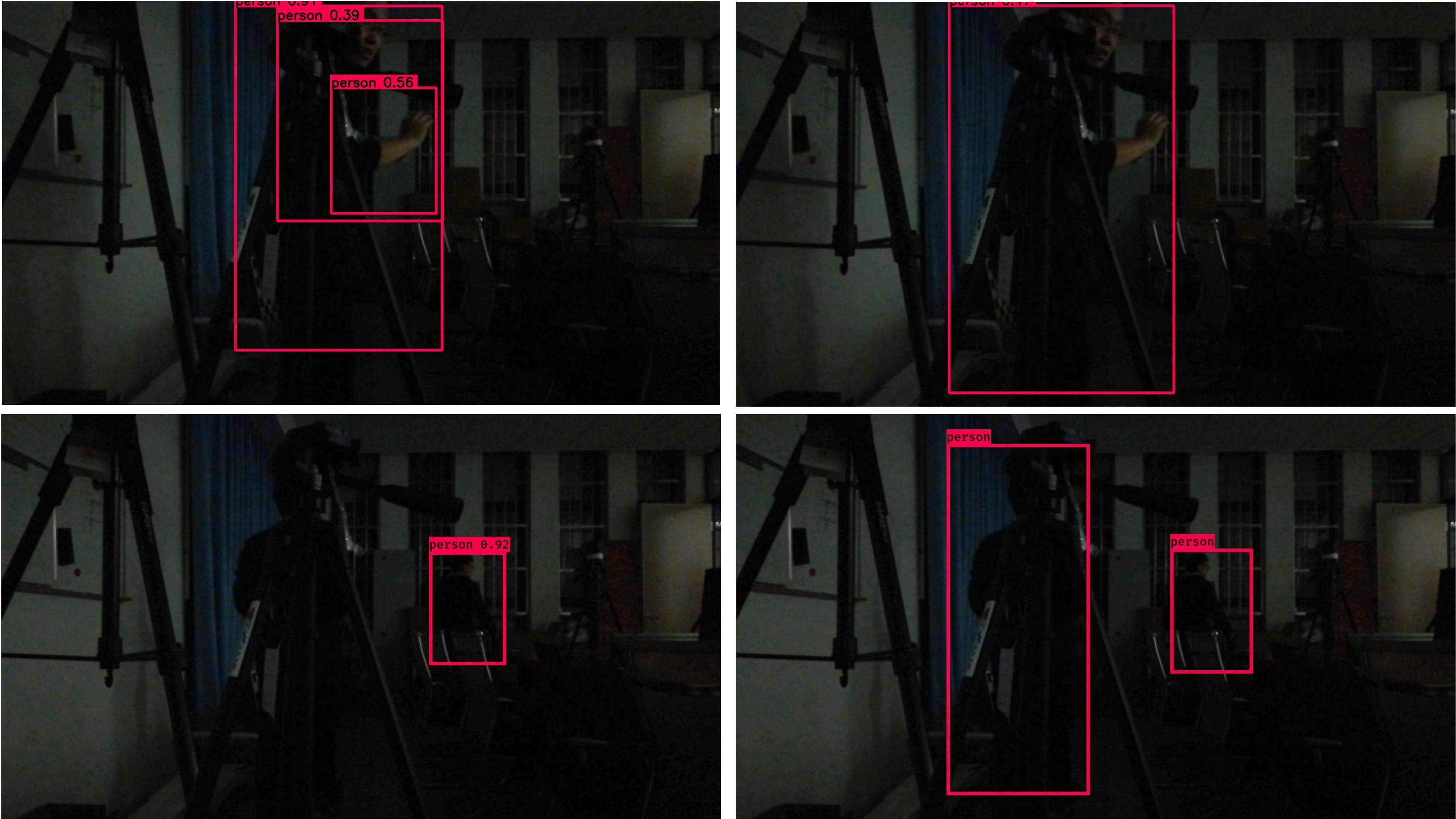}\\
  \caption{Detection results comparison of our dataset. The left column is the detection of the original detector. The right column is the detector with our proposed methods.}\label{butu}
\end{figure}

Figure \ref{butu} shows the performance comparison of our dataset. The above row shows the result comparison of mask R-CNN detector while the below row shows the result comparison of YOLO detector. Observing the above row, we can find it is the situation of false positive. The sum of the false positives and negatives in the left is 2 (two boxes on the same person) and the true detection ratio is $1/3$.
The below row stands for the situation of false negative.
The sum of the false positives and negatives in the left is 1 (one person's detection is missed) and the true detection ratio is $1/2$.
Both of our adopted evaluation methods will be influenced by redundant detection and missed detection without considering confidence score.
And with our proposed methods, we find the evaluation will be improved while the visual performance gets better as well.

Table \ref{OurDataset} shows the performance improvement of our methods on our dataset and the confidence thresholds of the original detection methods are both 0.3. We can find that with the localization and the proposed NMS, the false positives and false negatives can be restrained, while the ratio of the true positives can be increased.
Different from the results in simulated COCO dataset, we find that in this case the proposed method 1 performs better.
This is because in real world the localization precision may not be as accurate as the localization result simulated in COCO dataset.

\begin{figure*}[]
  \centering
  \includegraphics[width=1.0\linewidth]{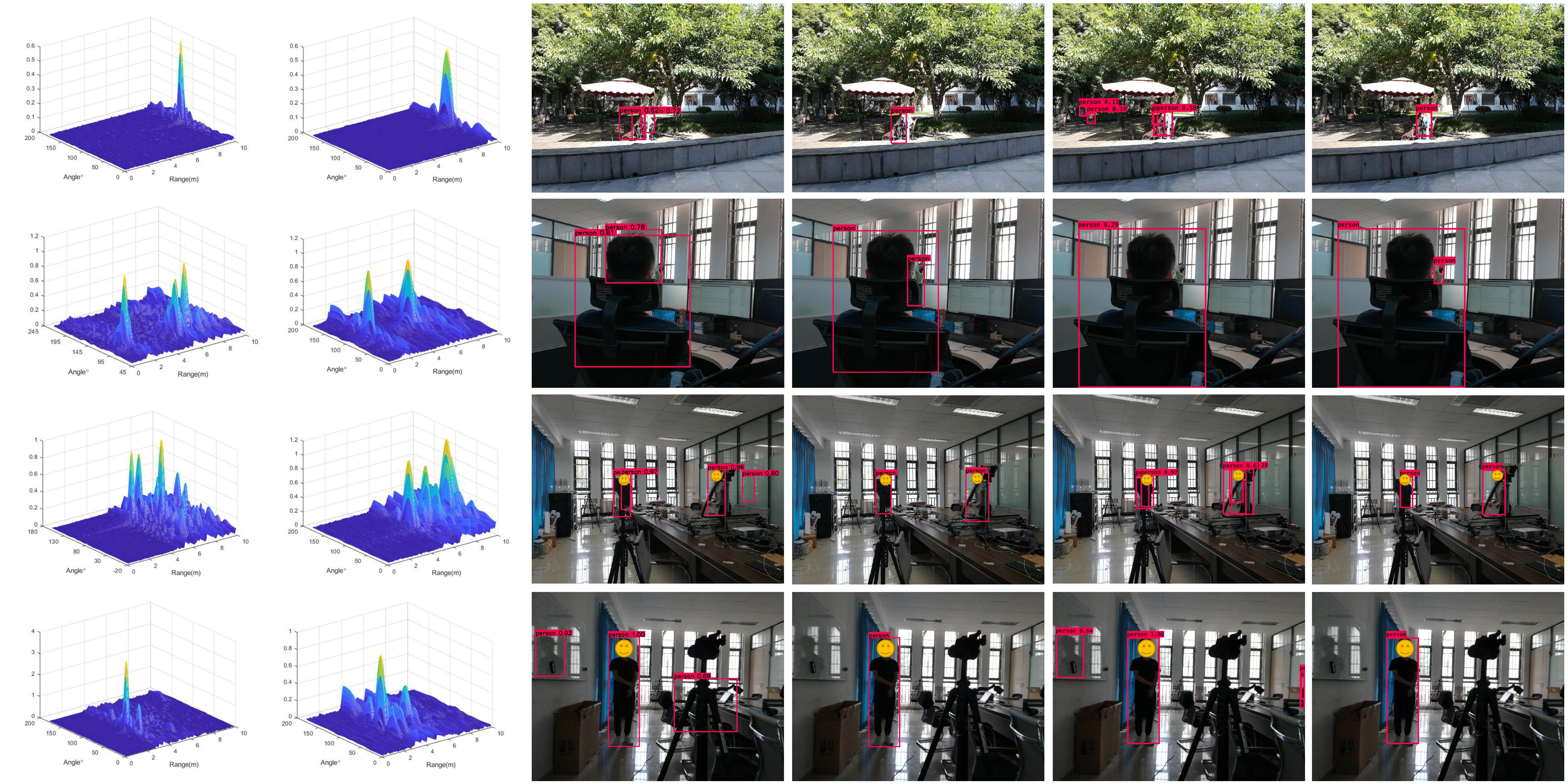}\\
  \caption{Detection results in real-world scenarios. 
  \textbf{First column:} Estimated horizontal AoA-ToF heatmap. In real-world environment, the heatmap would be disturbed by multi-path interference. Nevertheless, the final detection would not be affected since the AoA-ToF is only utilized as te initial estimate.   
  \textbf{Second column:} Estimated vertical AoA-ToF heatmap. Similar to the horizontal heatmap, the vertical heatmap is also disturbed by multi-path interference. \textbf{Third column:} Results of Mask R-CNN. \textbf{Fourth column:} Results of Mask R-CNN equipped with proposed method 2.  \textbf{Fifth column:} Results of YOLOv3. \textbf{Sixth column:} Results of YOLOv3 equipped with proposed method 1. }
  \label{Final}
\end{figure*}

Figure \ref{Final} illustrates more results in real-world scenarios. Similar to previous experiments, we conduct the real-world detection with the original Mask R-CNN detector, the original YOLOv3 detector, and our proposed detectors. For Mask R-CNN, we set the confidence threshold to 0.5 to make the detection results more visible. The peaks which are the potential object locations will be chosen if their magnitudes exceed half of the highest magnitude.
The results clearly demonstrate that with radio information, our proposed detectors can well address the problems of false positives and false negatives, compared with the original Mask R-CNN and YOLOv3.

\section{Conclusion}
In this paper, we proposed a human detection framework with the aid of radio information for anchor-based one-stage detector and two-stage detectors.
Systematically, based on the radio signals, we first estimated the localization of each person in the image with its angle and distance from the camera.
Then, we proposed two ways to utilize the radio localization information for anchor-based one-stage detector and two-stage detectors.
With the radio identifier and localization information, we also proposed a non-maximum suppression with an extra constraint that the radio localizations and the detections should be one-on-one matched. Experiments on simulative datasets and real-world scenarios showed that our proposed methods could improve the performance of state-of-the-art detectors and alleviate the problem of false positives and false negatives.

\quad

\bibliography{sample-base}
\bibliographystyle{IEEEtran}

\end{document}